\documentclass[twocolumn, twoside]{article}

\usepackage{arxiv}
\usepackage[flushleft]{threeparttable}

\usepackage[utf8]{inputenc} 
\usepackage[T1]{fontenc}    
\usepackage{hyperref}       
\hypersetup{
    colorlinks=true,
    citecolor=cyan,
    linkcolor=cyan,
    filecolor=cyan,      
    urlcolor=cyan,
    }
\usepackage{url}            
\usepackage{booktabs}       
\usepackage{amsfonts}       
\usepackage{nicefrac}       
\usepackage{microtype}      
\usepackage{cleveref}       
\usepackage{lipsum}         
\usepackage{graphicx}
\usepackage{doi}
\usepackage[round]{natbib}
\usepackage[switch]{lineno}

\title{Open Access Battle Damage Detection \\ via Pixel-Wise T-Test on Sentinel-1 Imagery}

\author{ {\hspace{1mm}Ollie Ballinger}\thanks{} \\
	Centre for Advanced Spatial Analysis\\
	University College London\\
	\texttt{o.ballinger@ucl.ac.uk} \\
}

\hypersetup{
pdftitle={Open Access Conflict Damage Detection},
pdfauthor={Ollie ~Ballinger},
}

\begin{document}

\twocolumn[ 
  \begin{@twocolumnfalse} 
  
\maketitle

\begin{abstract}
In the context of recent, highly destructive conflicts in Gaza and Ukraine, reliable estimates of building damage are essential for an informed public discourse, human rights monitoring, and humanitarian aid provision. Given the contentious nature of conflict damage assessment, these estimates must be fully reproducible, explainable, and derived from open access data. This paper introduces a new method for building damage detection-- the Pixel-Wise T-Test (PWTT)-- that satisfies these conditions. Using a combination of freely-available synthetic aperture radar imagery and statistical change detection, the PWTT generates accurate conflict damage estimates across a wide area at regular time intervals. Accuracy is assessed using an original dataset of over half a million labeled building footprints spanning 12 cities across Ukraine, Palestine, Syria, and Iraq. Despite being simple and lightweight, the algorithm achieves building-level accuracy statistics (AUC=0.88 across Ukraine, 0.81 in Gaza) rivalling state of the art methods that use deep learning and high resolution imagery. The workflow is open source and deployed entirely within the Google Earth Engine environment, allowing for the generation of interactive Battle Damage Dashboards for \href{https://ollielballinger.users.earthengine.app/view/ukraine-damage-assessment}{Ukraine} and \href{https://ee-ollielballinger.projects.earthengine.app/view/gazadamage}{Gaza} that update in near-real time, allowing the public and humanitarian practitioners to immediately get estimates of damaged buildings in a given area.
\end{abstract}

\vspace{0.35cm}

  \end{@twocolumnfalse} 
] 

\section{Introduction}
The accurate and timely identification of buildings damaged by war is crucial for local civilians, aid agencies, and an informed public. Recent progress has been made in automatically identifying damaged buildings using deep learning and very-high resolution (VHR) optical satellite imagery. However, these models often struggle to generalize and are both financially and computationally expensive. As such, there has been little uptake of these methods by humanitarian practitioners such as the United Nations Satellite Agency (UNOSAT), which continues to identify damaged buildings by manually combing through satellite imagery. 

This paper develops a new algorithm for the detection of urban conflict damage-- the Pixelwise T-Test (PWTT)-- that is lightweight, unsupervised, and uses only freely-available Synthetic Aperture Radar (SAR) imagery. This addresses many of the problems associated with expense, coverage consistency, explainability, and domain shift that are inherent to the use of deep learning on optical imagery. 

The recent, devastating conflicts in Gaza and Ukraine have evidenced significant public demand for open access conflict damage estimates; the PWTT was deployed by the Economist magazine to assess building damage across all of Ukraine at multiple time intervals in the first year of the conflict \citep{economist_data_2023}. Similarly, most major international media organizations have published SAR-derived building damage estimates to quantify the destruction wrought by the Israeli bombardment of the Gaza strip \citep{shah_ruined_2023, leatherby_maps_2023}. Yet, there are no studies that conduct large scale, building-level accuracy assessments of SAR-derived conflict damage detection. 

As such, a second key contribution of this paper is the creation and release of a novel dataset of 633,199 manually labeled building footprints spanning 12 cities in Ukraine, Palestine, Syria, and Iraq. This is used to conduct comprehensive validation of the PWTT algorithm in different conflict settings. Results indicate that the PWTT can identify damaged buildings with accuracy rivalling that of deep learning approaches, achieving an Area Under the Curve of 0.88 across 8 Ukrainian cities and 0.81 for the entire Gaza strip.

Finally, the lightweight and open-access nature of the PWTT enables the creation of dynamic Battle Damage Dashboards, updating with new imagery in near-real time, offering country-wide damage assessments across any war time frame. The dashboards integrate damage estimates with high-resolution pre-war population data, gauging the displaced population in destroyed areas. The dashboards incorporate other open-source data, such as geolocated social media footage, providing supplementary verification. This streamlined, adaptable approach signifies a significant advancement in real-time conflict impact analysis.

\subsection{Building Damage Assessment using Deep Learning}

Considerable progress has been made in the field of building damage assessment (BDA) using remote sensing. One of the most significant recent developments has been release of the xBD dataset, which consists of over 45,000 $km^2$ of high resolution satellite imagery and over 850,000 annotated building footprints from natural disasters around the world \citep{gupta_xbd_2019}. xBD has become the benchmark dataset for both training and evaluating deep learning-based building damage detection models, which have proliferated since \citep{shen_bdanet_2022, xia_deep_2023, deng_post-disaster_2022}. However, no equivalent benchmark exists for conflict-induced building damage assessment and there are several domain-specific challenges that distinguish the latter from general building damage assessment from natural disasters. 

Whereas natural disasters typically involve a single destructive event affecting a relatively small area, battle damage can gradually accrue over months or years across a very large area. Neural networks work well for the former, but are harder to implement for multi-temporal wide-area monitoring. One of the most influential recent studies of conflict-induced destruction is Mueller et. al.'s (2021) study of the Syrian civil war, which achieves high accuracy by training a Convolutional Neural Network on damage annotations carried out by the United Nations and high resolution satellite imagery \citep{mueller_monitoring_2021}. When training and testing on the same cities, they achieve an Area Under the Curve (AUC) of 0.9. Yet the model struggles to generalize even within Syria: when a model is only trained on Aleppo and inference is conducted on other Syrian cities, accuracy statistics drop by as much as 29\%, with the  average AUC for these cities falling to 0.7. 

Compared to natural disasters, the contentious nature of battle damage assessments impose a set of additional requirements for verifiable impartiality. Results must be fully reproducible-- requiring both code and necessary inputs to be open access-- and explainable. Neural networks are often referred to as "black boxes", and do not yield explainable results. Most deep learning approaches also eschew open data in favor of expensive, high resolution optical imagery; monitoring large areas can become very costly, particularly if repeated observations are required. At \$23/$km^2$ for new high-resolution satellite imagery from MAXAR (the kind used by Mueller et. al. (2021), and in the xBD dataset), it would cost \$13.9 million to image the entirety of Ukraine just once \citep{mueller_monitoring_2021, gupta_xbd_2019, aaas_high-resolution_2024}. Clouds pose an additional challenge for optical imagery; while damage assessment for conflicts in the Middle East is not significantly hampered by cloud cover, certain areas in Ukraine are often shrouded in clouds. As such, while much of the recent literature on building damage detection utilizes high resolution optical imagery and deep learning, these methods are less well suited to the study of conflict-induced damage. 

\begin{figure*}[h!]
  \caption{Time-series Sentinel-1 Backscatter Amplitude of a Damaged Building in Mariupol, Ukraine}
      \begin{minipage}{\textwidth}
  \includegraphics[width=\textwidth]{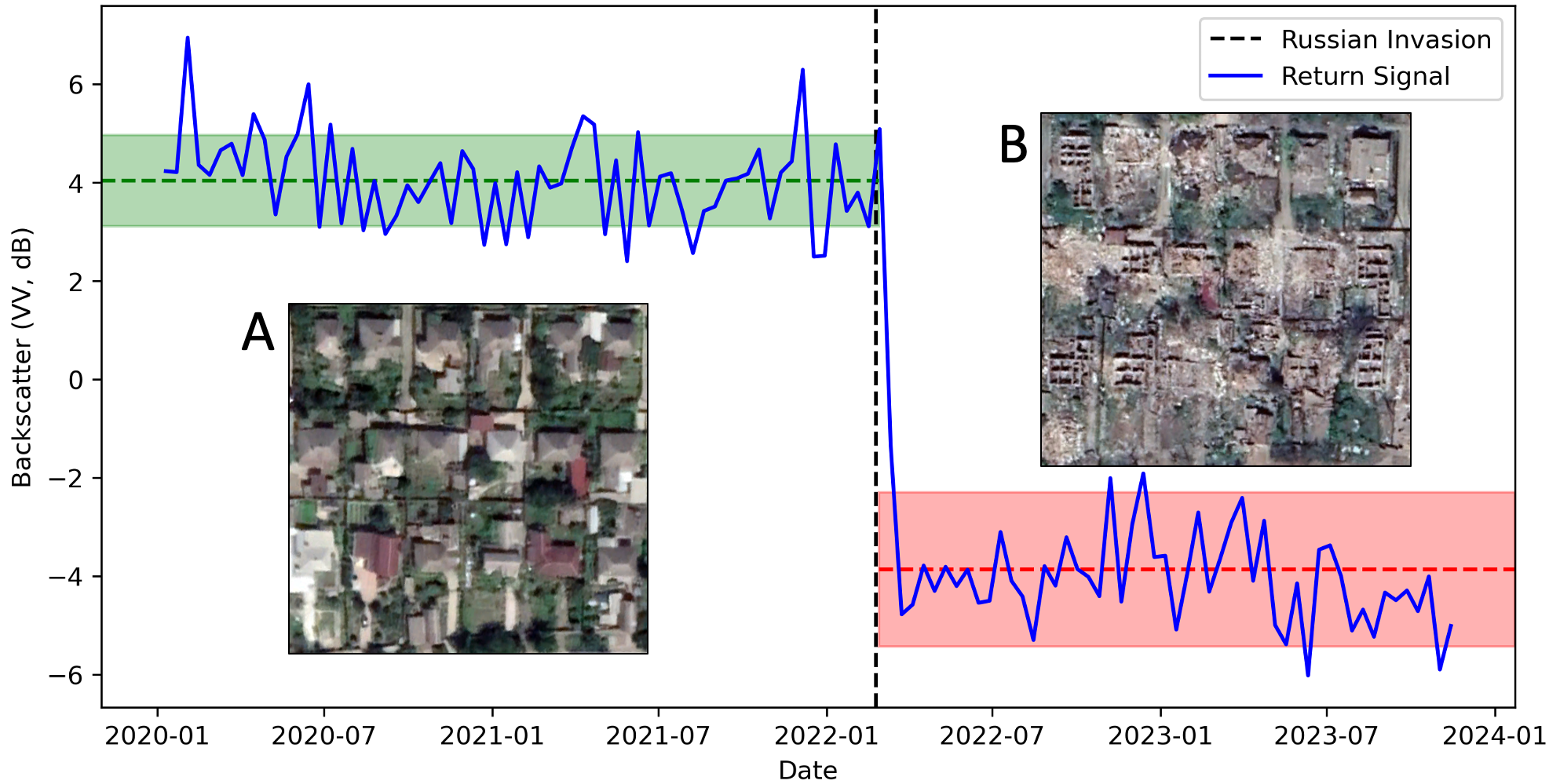}
{\scriptsize Change in backscatter amplitude resulting from the destruction of a building in Mariupol, Ukraine, following the Russian invasion. The green dashed line and shaded area represent the pixel's mean backscatter amplitude $\pm$ 1 standard deviation prior to the invasion, while the red line and shaded area represent these statistics following the building's destruction. Subfigures A and B show optical imagery of the area before and after destruction. \par}
\end{minipage}
  \label{pixel_ts}
\end{figure*}

\subsection{Building Damage Assessment in Synthetic Aperture Radar Imagery}

Synthetic Aperture Radar (SAR) imagery has a number of properties that makes it useful for building damage assessment, including weather- and illumination independent imaging and a high repeat time \citep{esa_sentinel-1_2024}. The SAR sensor emits a pulse of microwaves towards the earth and measures the return signal (backscatter), enabling the analysis of textural features on the earth’s surface based on how they reflect the signal back to the satellite.

There is a vast literature on building damage detection using SAR imagery, largely in the field of natural disaster response. These are broadly divided into two categories, on the basis of whether they rely on detecting changes in the phase or the amplitude of the return signal. The approach taken herein conducts amplitude-based change detection, leveraging the unique characteristics of conflict-induced destruction as well as advances in computing power to accurately identify battle damage.

The most common amplitude-based approach to building damage detection involves simply calculating the difference or ratio between one pre-disaster image and one post-disaster image \citep{liu_urban_2011, gokon_towards_2017, dellacqua_mapping_2010, wang_statistical_2009}. A significant change in signal amplitude would have to result from a change in a building’s scattering mechanism-- for example the transition from double-bounce scattering to diffuse scattering as a standing building turns to rubble-- rather than smaller changes \citep{ge_review_2020}. However, backscatter amplitude is affected by the dielectric properties of the target, such as the roofing material and the presence of water or snow which may reduce accuracy particularly if only a small number of images are used. Several studies have found that given only a pair of images, phase-based change detection outperforms amplitude-based methods particularly for the detection of minor damage \citep{liu_extraction_2017, arciniegas_coherence-_2007}

As such, coherence change detection (CCD) is the more widely used method for building damage assessment in SAR imagery, and this approach has a number of advantages. Because it analyzes the shift in the returning signal’s phase, CCD is able to pick up on minute changes on the ground owing to the fact that Sentinel-1 transmits C-band signals with a wavelength of just 5.6 centimeters \citep{ge_review_2020}. This sensitivity has made CCD particularly widespread in the analysis of earthquake damage, as it is not only capable of detecting damaged buildings but centimeter-level ground subsidence \citep{ito_extraction_2000, yonezawa_decorrelation_2001}. However, this sensitivity can also be a liability, as “coherent change detection methods tend to have very large false positive rates, where change is vastly overestimated.” \cite[p.~19]{meyer_spaceborne_2019}. Furthermore, the image processing involved in the calculation of CCD results in a 16-fold reduction in resolution, from 10 meters per pixel to 40m/px. At this resolution, 98\% of the building footprints in the validation dataset are smaller than an individual pixel. High sensitivity and low resolution, particularly in the context of conflict, create a significant risk of false positives. 

\section{Methods}

The damage detection algorithm developed herein is a multi-temporal Pixel-Wise T-Test computed using backscatter amplitude, which mitigates many of the limitations of existing coherence and amplitude-based building damage detection methods. By relying on backscatter amplitude, it allows for the maintenance of the imagery's native resolution of 10 meters and is not overly sensitive to minor changes when compared with coherence change detection. By integrating pre-and post-conflict pixel standard deviation over extended time periods, it better distinguishes building damage from general anthropogenic change and seasonal variation that often affect amplitude-based methods. The computation of the test statistic is simple enough to be calculated by hand, and can consequently be applied over very large areas at regular time intervals. The use of time-series statistical change detection obviates the need for training and fine-tuning, and mitigates many of the problems associated with overfitting and domain shift that plague neural networks. 

\subsection{Data}

\subsubsection{Synthetic Aperture Radar Imagery}

The main source of data used in this study is Synthetic Aperture Radar (SAR) imagery from the European Space Agency’s Sentinel-1 satellite. Sentinel-1 is a dual polarimetric SAR system, emitting C-band radar microwaves with a vertical polarization and recording both vertical and horizontal reflected polarizations \citep{esa_sentinel-1_2024}. Two polarizations are used in this analysis: single co-polarization (vertical transmit/vertical receive, or VV) and dual-band cross-polarization (vertical transmit/horizontal receive, or VH). These polarizations are differentially sensitive to different types of scattering mechanisms generated by ground targets, such as double-bounce scattering generated by standing buildings or rough surface scattering generated by rubble \citep{meyer_spaceborne_2019, canada_radar_2008}. Previous studies have found that a combination of VV and VH polarizations improved building damage detection accuracy \citep{karimzadeh_sadra_building_2017}.

Ground Range Detected (GRD) scenes acquired in Interferometric Wide-Swath mode are accessed in Google Earth Engine at a spatial resolution of 10 meters per pixel \citep{esa_sentinel-1_2024}. The collection is updated in near real time, and undergoes orthorectification, thermal noise removal, and radiometric calibration prior to its ingestion into the Earth Engine catalog \citep{esa_sentinel-1_2024}. An additional pre-processing step-- speckle filtering-- is conducted prior to analysis using a Lee Filter \citep{lee_speckle_1994}. 

\subsubsection{Annotated Building Footprints}

Accuracy assessments are carried out using an original dataset of 633,199 annotated building footprints, spanning 12 cities in four different countries. This dataset was compiled by spatially joining damage annotations from the United Nations Satellite Centre (UNOSAT) with data on building footprints. This allows for the generation of more meaningful accuracy statistics, and a more stringent test of model accuracy than previous studies. Accuracy assessment at the building level follows the standards set by xBD dataset \citep{gupta_xbd_2019}.

Building footprint data is sourced from the Microsoft Building Footprints dataset, which consists of over 1 billion building footprints derived from high resolution satellite imagery around the world using deep learning \citep{microsoft_microsoftglobalmlbuildingfootprints_2024}. For the Middle East and Europe, buildings are detected with a precision of 95\% and recall of 85\%. A manual review of the footprints found them to be generally high quality, but that many of the very small (<$50m^2$) building footprints appeared to be false positives (cars/trucks, garden sheds, or errant geometries), and were removed. 

A building footprint is labeled as damaged if it intersects with a UNOSAT damage annotation point, and labeled undamaged otherwise. UNOSAT annotations are generated manually on the basis of high resolution optical satellite imagery \citep{unosat_irpin_2022, unosat_chernihiv_2022, unosat_damage_2016, unosat_damage_2017, unosat_damage_2017-1, unosat_hostomel_2022, unosat_kharkiv_2022, unosat_lysychansk_2022, unosat_mariupol_2022, unosat_rubizhne_2022, unosat_sievierodonetsk_2022, unosat_unosat_2024}. These annotations are the most commonly used source of labeled data in studies of battle damage, and are generally regarded to be of high quality \citep{mueller_monitoring_2021,kahraman_battle_2016, witmer_remote_2015, boloorani_post-war_2021, braun_assessment_2018, huang_monitoring_2023}. However, because the imagery used for annotation is on-nadir, some damage caused to the sides of buildings may be missed. 

\begin{table}[h!]
\begin{threeparttable}
\caption{Validation Dataset}
\begin{center}
\tabcolsep=0.15cm
\begin{tabular}{p{0.8cm}p{1.5cm}p{0.5cm}p{0.5cm}p{1.7cm}}
\toprule
\midrule
\multicolumn{1}{l}{{Country}} & \multicolumn{1}{l}{{City}} &\multicolumn{1}{l}{{Footprints}}& \multicolumn{1}{c}{{\shortstack[c]{Percent \\ Damaged}}}& \multicolumn{1}{c}{\shortstack[c]{Annotation \\ Date}}\\
\midrule
Palestine & Gaza & 201629 & 33.88 & 2024-01-07 \\
Ukraine & Lysychansk & 20246 & 7.42 & 2022-09-21 \\
Ukraine & S'odonetsk & 5970 & 24.32 & 2022-07-27 \\
Ukraine & Rubizhne & 8899 & 33.67 & 2022-07-09 \\
Ukraine & Kharkiv & 107976 & 0.85 & 2022-06-15 \\
Ukraine & Mariupol & 18446 & 31.42 & 2022-05-12 \\
Ukraine & Hostomel & 4463 & 13.85 & 2022-03-31 \\
Ukraine & Irpin & 7288 & 11.21 & 2022-03-31 \\
Ukraine & Chernihiv & 29929 & 3.25 & 2022-03-22 \\
Syria & Raqqa & 24689 & 45.36 & 2017-10-21 \\
Iraq & Mosul & 137794 & 11.51 & 2017-08-04 \\
Syria & Aleppo & 65870 & 30.82 & 2016-09-18 \\
\midrule
\bottomrule
\end{tabular}

\begin{tablenotes}
\scriptsize
\item
By joining manual damage annotations from UNOSAT with building footprints using a 10 meter tolerance, 633,199 labeled building footprints spread across 12 cities in 4 different conflict zones are used for validation. 
\end{tablenotes}
\end{center}
\end{threeparttable}

\label{val_tab}
\end{table}

Twelve cities were selected to assess the accuracy of the proposed methodology, on the basis of four criteria: the availability of Sentinel-1 imagery, reliable building footprints data, UNOSAT damage annotations, and a minimum proportion of damaged buildings (\>0.5\%). The resulting dataset has even coverage across four different conflict zones, with 32\% of the footprints located in the Gaza Strip, 32\% in Ukraine, 22\% in Iraq, and 14\% in Syria. These cities encompass a wide range of different climate and atmospheric conditions, building types, urban morphologies, and damage levels, ranging from just 0.8\% damage in Kharkiv to 45\% damage in Raqqa. Table \ref{val_tab} provides summary statistics for the verification dataset. 

\subsection{The Pixel-Wise T-Test}

Figure \ref{pixel_ts} demonstrates the change in backscatter amplitude for a destroyed building in Mariupol, Ukraine, following its destruction. The corresponding Sentinel-1 pixel has a low standard deviation in both the pre-and post-war periods, but experiences a large change in mean amplitude. The T-Test is a simple signal-to-noise ratio that measures the difference between the means of two samples adjusted by the standard deviation within each sample. The integration of multi-temporal pixel standard deviation into the test allows for better isolation of damage, rather than general anthropogenic change as detected by bi-temporal algorithms \citep{canty_mortimer_statistical_2019}; Airports and train stations are likely to have high pixel variance over time, meaning that these areas are unlikely to achieve high T-values. As such the T-test is well suited to the detection of building damage resulting from conflict.

The first step in the calculation of the test statistic is the selection of a pre-war reference period ($\tau=0$) and a post war inference period ($\tau=1$) using a cutoff date (e.g. the beginning of hostilities in a given area). Figure \ref{acled} in the appendix shows the selection of reference and inference periods for the cities in the sample along with the weekly number of clashes therein. In all cases, the start of the inference period corresponds to the date of the UNOSAT damage assessment used for validation, and spans two months thereafter. The reference periods for Gaza and all Ukrainian cities span one year prior to the onset of hostilities (10/10/2024 and 24/02/2022, respectively). For Mosul, Raqqa, and Aleppo, the reference periods avoid the peak of the fighting. 

The pre-war reference period spans a full year ($\sim$30 images) in order to capture the effect of seasonal changes in the dielectric properties of the target (e.g., the presence of snow). The inference period uses two months' worth of imagery ($\sim$5 images)-- shorter time periods result in too small a sample for inference, and the use of longer time periods increases the risk of damage accruing during the inference period. For optimal performance, there should be minimal change during both sample periods. This is the case for Ukraine and Gaza which both featured an abrupt onset of hostilities. However, the algorithm still performs well when there is substantial damage occurring during the reference period, as is the case for the cities in Syria and Iraq. 

Following the selection of a pre-conflict reference period and a post-conflict inference period, the calculation of pixel-level t-values proceeds as follows. The mean backscatter amplitude $\overline{x}$ for each pixel $x$ is calculated for all unique combinations of orbital pass $\omega$, polarization $\pi$, and time period $\tau$. Variable $n$ denotes the number of Sentinel-1 scenes.

\begin{equation}
\overline{x}_{\omega\pi\tau}=\frac{1}{n_{\omega\pi\tau}
}\sum_{i=1}^{n_{\omega\pi\tau}} x_{\omega\pi\tau}    
\end{equation}

The corresponding pixel-level standard deviation $s$ is also calculated.

\begin{equation}
s_{\omega\pi\tau}=\sqrt{\frac{1}{n_{\omega\pi\tau}-1}\sum_{i=1}^{n_{\omega\pi\tau} } (x_{\omega\pi\tau} - \overline{x}_{\omega\pi\tau})}
\end{equation}

A standard two-tailed T-test is then computed for each combination of orbital trajectory (ascending/descending) and polarization (VV/VH).

\begin{equation}
t_{\omega\pi} = {\frac{\overline{x}_{\omega\pi\tau_0}-\overline{x}_{\omega\pi\tau_1}}{\sqrt{\frac{s^2_{_{\omega\pi\tau_0}}}{n_{_{\omega\pi\tau_0}}} + \frac{s^2_{_{\omega\pi\tau_1}}}{n_{_{\omega\pi\tau_1}}}}}} 
\end{equation}

There are thus four t-values per pixel. The final change statistic $T$ is the maximum of the absolute t-values, because the directionality of the change in backscatter resulting from the destruction of a building is unknown a-priori. 

\begin{equation}
T=\max\{|t_{\omega \pi}|\}
\end{equation}

\begin{figure*}[!h]
  \caption{Damage Prediction and Accuracy Assessment, Mosul}
        \begin{minipage}{0.53\textwidth}
             \includegraphics[width=\columnwidth]{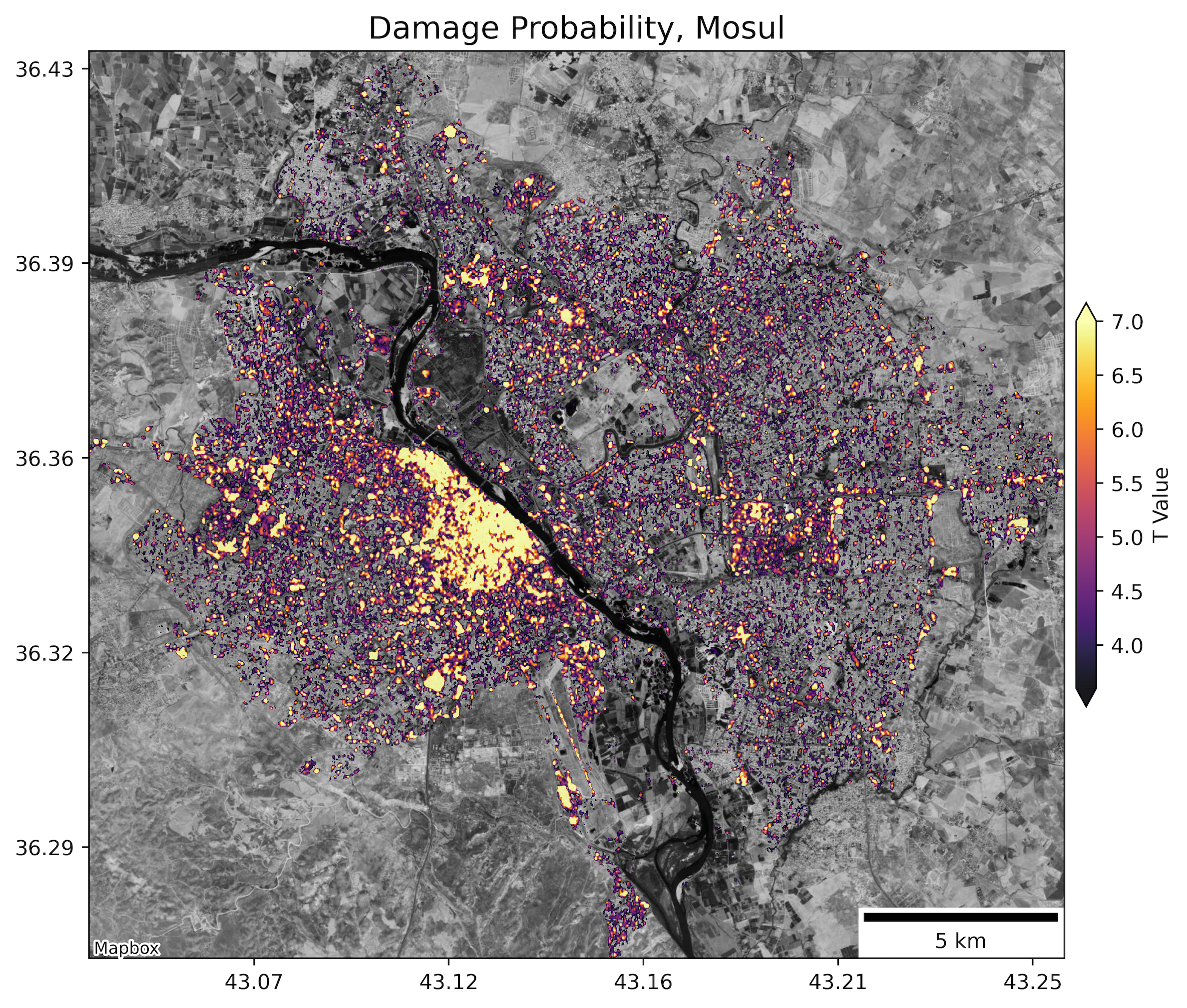}
        \end{minipage}
        \begin{minipage}{0.47\textwidth}
             \includegraphics[width=\columnwidth]{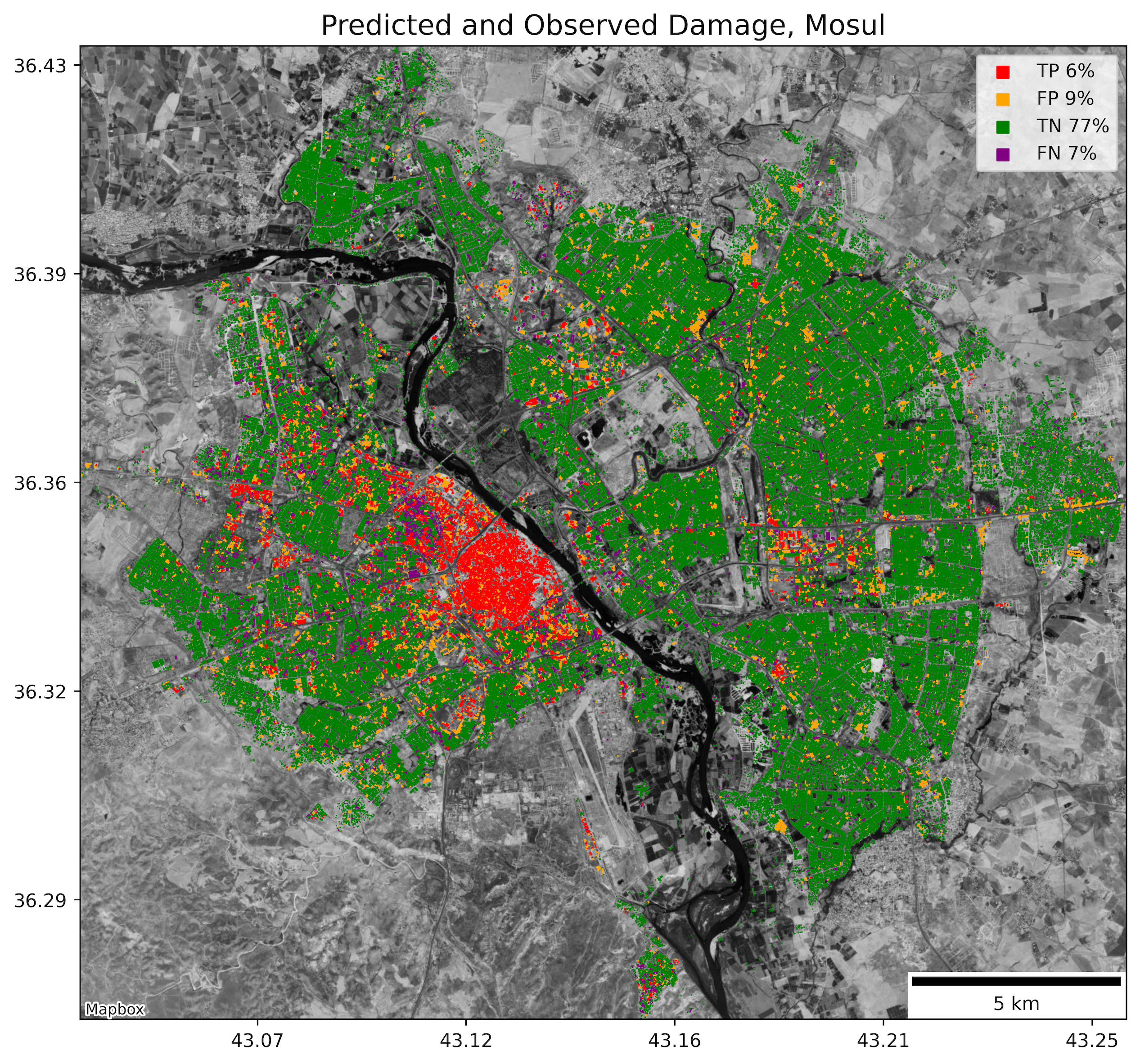}
        \end{minipage}
{\scriptsize The damage probability raster for Mosul is shown on the left. When joined with the labeled building footprints (right), True Positive and True Negative predictions are shown in red and green, respectively, while False Positives and False Negatives are shown in orange and purple. Percentages in the legend reflect the proportion of building area in each category. Though Mosul has the second-lowest F1 accuracy score (0.42) out of the entire sample, the PWTT still correctly localizes damage in the city.

\par}
  \label{mosul}
\end{figure*}

Assuming that damage generally increases surface roughness, depending on a building’s geometry and orientation relative to the satellite, the transition from a standing structure to rubble could either lead to an increase or decrease in backscatter amplitude. For example, large structures with flat roofs such as warehouses or shopping centers might experience an increase in backscatter amplitude after being damaged as the scattering mechanism transitions from specular reflection to diffuse scattering \citep{huang_monitoring_2023}. Conversely, a free-standing house that previously generated double-bounce scattering would experience a drop in backscatter amplitude after becoming destroyed, as is likely the case in Figure \ref{pixel_ts}. 


\subsection{Damage Classification}

After the generation of the damage probability raster, building-level inference is carried out by calculating the mean pixel value within each building footprint, and setting a threshold value for binary damage classification (damaged/undamaged).

A key advantage of the PWTT is that the damage probability raster contains non-arbitrary values. A threshold value of $T$ can be chosen according to the desired statistical significance (e.g. $T>2.7$ where $n=40$ at the 99\% confidence level). This approach classifies statistically significant changes that occurred during the post-conflict period as damage. T-values also generalize across domains, since the T-test is a measure of relative change and accounts for absolute differences in backscatter values across geographic areas. The threshold value of T can also be selected empirically, using a precision-recall curve (as shown in Figure \ref{roc}), allowing for the selection of a more context-aware threshold for a given geography.

\section{Results}

\subsection{Building-Level Damage}

Damage predictions are compared to a validation dataset of 633,199 labeled building footprints. Figure \ref{mosul} demonstrates the process of accuracy assessment in Mosul. The damage probability raster on the left predicts heavy damage to the Old City on the south bank of the Tigris River. The plot on the right joins the damage predictions with the labeled building footprints, showing True Positives in red, True Negatives in green, False Positives in orange, and False Negatives in purple. The PWTT accurately localizes damage, not just in the Old City but in isolated pockets to the West as well. This process is carried out for all 12 cities (corresponding plots available in the Appendix), and building-level accuracy metrics are reported in Table \ref{results_tab}.

\begin{table}[b!]
\begin{center}

\begin{threeparttable}
\caption{Building-Level Damage Detection Accuracy}

\begin{tabular}{llllll}
\toprule
\midrule
City & AUC & F1 & Precision & Recall & N \\
\midrule
Gaza & 0.81 & 0.64 & 0.53 & 0.82 & 201629 \\
Lysychansk & 0.77 & 0.49 & 0.5 & 0.48 & 20246 \\
S'odonetsk & 0.72 & 0.64 & 0.49 & 0.92 & 5970 \\
Rubizhne & 0.8 & 0.69 & 0.66 & 0.73 & 8899 \\
Kharkiv & 0.82 & 0.28 & 0.25 & 0.32 & 107976 \\
Mariupol & 0.72 & 0.7 & 0.59 & 0.86 & 18446 \\
Irpin & 0.87 & 0.6 & 0.52 & 0.71 & 7288 \\
Hostomel & 0.85 & 0.68 & 0.63 & 0.74 & 4463 \\
Chernihiv & 0.88 & 0.5 & 0.46 & 0.55 & 29929 \\
Raqqa & 0.77 & 0.75 & 0.64 & 0.91 & 24689 \\
Mosul & 0.76 & 0.42 & 0.39 & 0.46 & 137794 \\
Aleppo & 0.7 & 0.6 & 0.48 & 0.8 & 65870 \\
\midrule
All & 0.78 & 0.52 & 0.4 & 0.73 & 633199 \\
\midrule
\bottomrule
\end{tabular}
\label{results_tab}
\begin{tablenotes}
\scriptsize
\item
Building-level accuracy statistics are generated using UNOSAT damage annotations joined with building footprints. For each location, a threshold value of T is selected using a precision-recall curve to generate a binary damage classification. All statistics are weighted by building area. 
\end{tablenotes}
\end{threeparttable}
\end{center}
\end{table}

\begin{figure*}[!h]
  \caption{Country-level ROC and Precision-Recall curves}
      \begin{minipage}{\textwidth}
  \includegraphics[width=\textwidth]{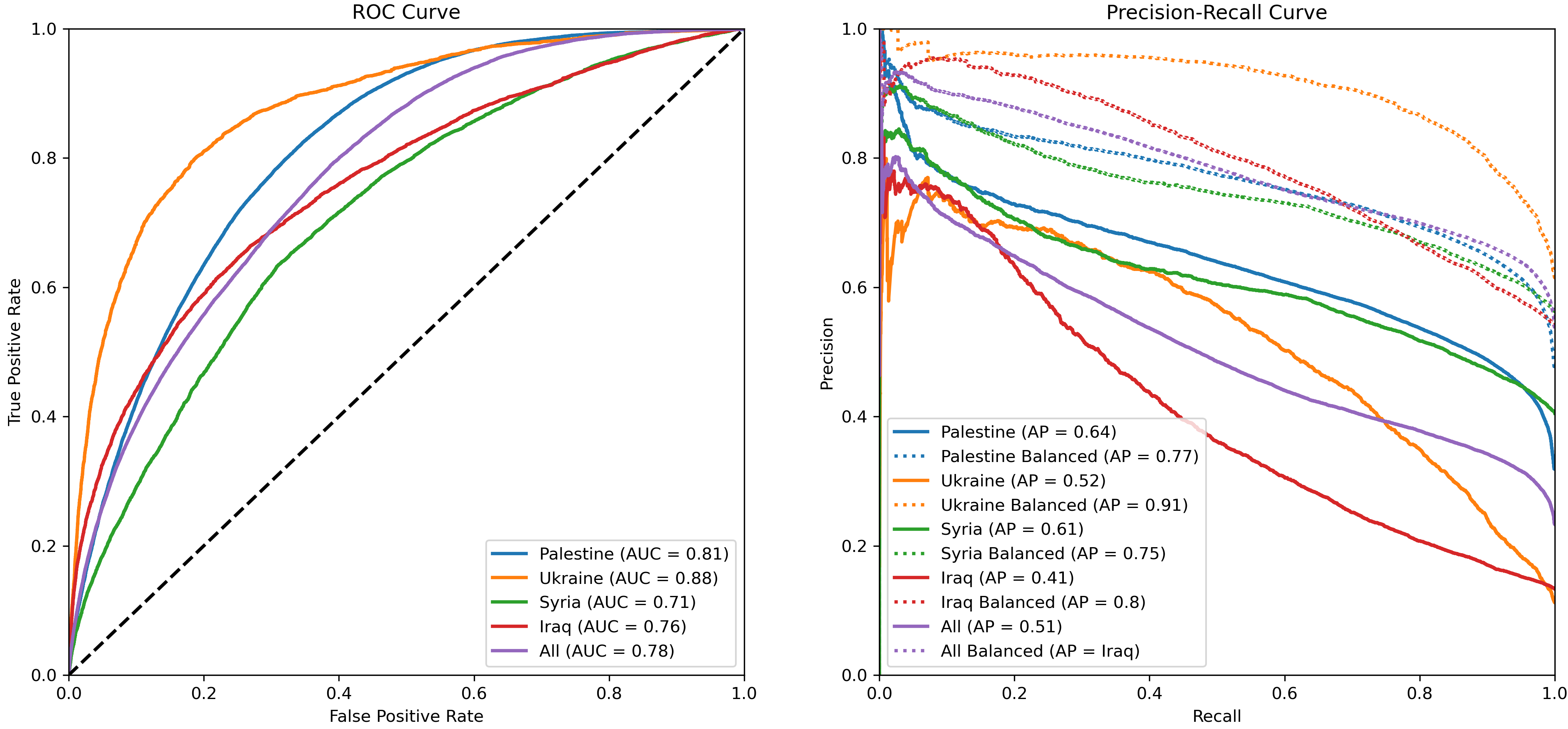}
{\scriptsize 
The Receiver Operating Characteristic (ROC) curve plots the true positive and false positive rates at each threshold value of T. The Precision-Recall curves does the same with precision and recall. Balanced samples were generated by randomly sampling the same number of undamaged buildings as damaged buildings for each country. City-level curves are available in Figure \ref{roc_city} in the Appendix. 
\par}
\end{minipage}
  \label{roc}
\end{figure*}

Accuracy-- the ratio of correct predictions to the total-- is the simplest gauge of model performance, but is strongly affected by class imbalance; in a city where only 1\% of buildings are damaged, a model blindly predicting everything to be undamaged would still garner 99\% accuracy.

By ignoring true negatives, the F1 score is more robust to class imbalance and is a more widely reported accuracy statistic in the study of building damage. It consists of the harmonic mean of Precision (the proportion of positive predictions that are correct) and Recall (the proportion of positive labels that are retrieved): 

$$
Precision = \frac{TP}{TP + FP} 
$$
$$
Recall  = \frac{TP}{TP + FN} 
$$
$$
F_1 = 2 \cdot \frac{Precision \cdot Recall}{Precision + Recall}
$$

Using a single threshold value of T for all of the building footprints in the sample, the PWTT achieves an F1 score of 0.52. This outperforms the baseline deep learning model trained with the release of the xBD dataset, which produced a building-level F1 score of 0.46 for the detection of buildings destroyed by natural disasters \citep{gupta_xbd_2019}. Yet, while the latter required over 45,000 km2 of expensive high resolution satellite imagery and 8 GPUs to train the model, the PWTT is unsupervised and employs satellite imagery that is freely available and nearly 400 times lower resolution. The highest F1 scores are achieved in Raqqa (0.75) and Rubizhne (0.69), while the lowest are observed in Mosul (0.42) and Kharkiv (0.25). 

Though less biased than simple accuracy, the F1 score is still sensitive to class imbalance. When comparing these results to the percentage of damaged buildings in a given city, it becomes apparent that F1 scores are positively correlated with the extent of destruction in a city: 45\% of buildings in Raqqa are damaged, compared with less than 1\% of buildings in Kharkiv.  As such, comparing F1 scores across cities can be misleading: given a classifier with a non-zero false positive rate, F1 approaches zero with the addition of new observations of the negative class. When a balanced sample (with the same number of damaged and undamaged buildings) is used, the F1 score for all of Ukraine increases to 0.85, and to 0.76 for Gaza (see Appendix, Table \ref{bal_results_tab}). 

\begin{figure*}[!h]
  \caption{Predicted and Observed Damage, Rubizhne}
      \begin{minipage}{\textwidth}
  \includegraphics[width=\textwidth]{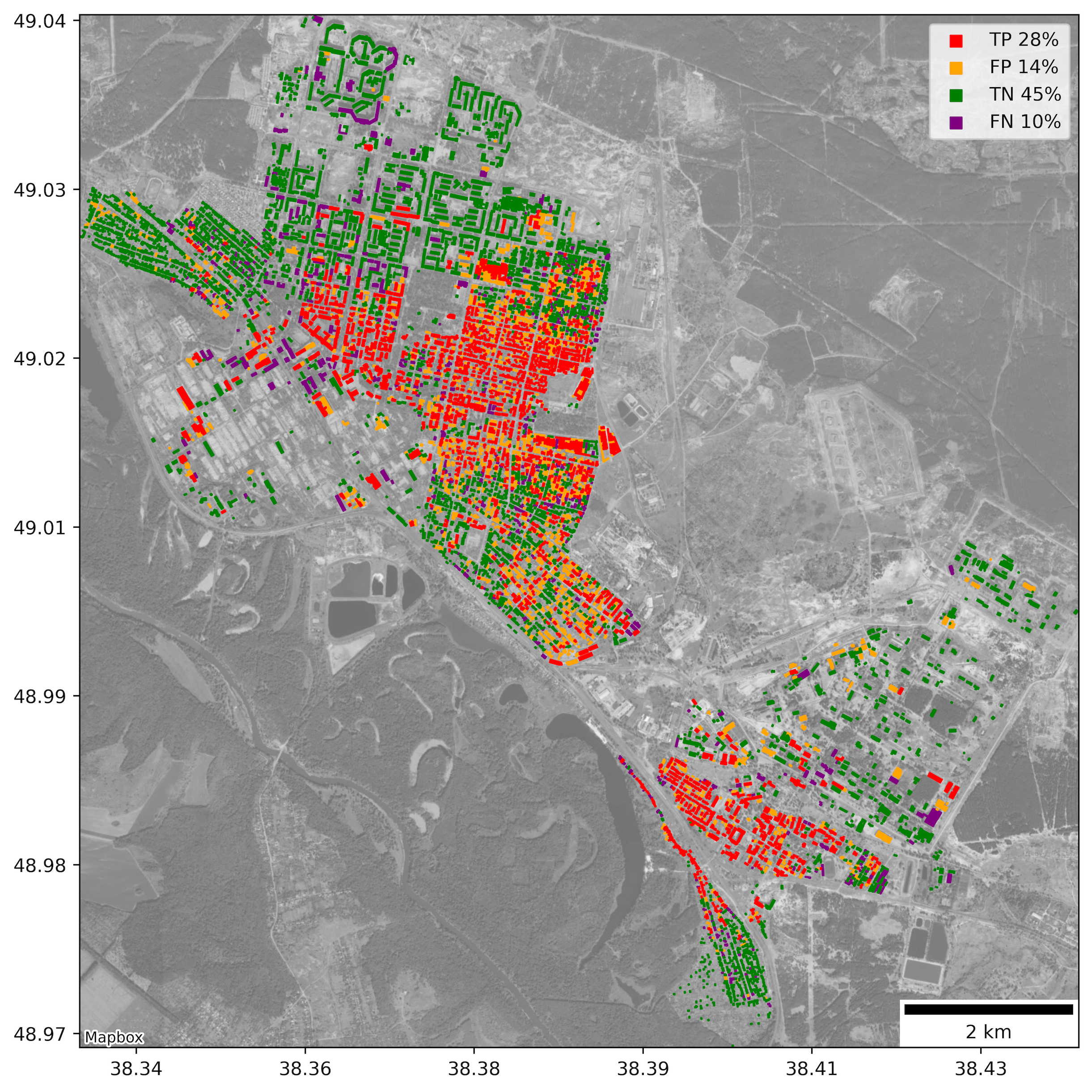}
{\scriptsize 
Predicted damage is compared to observed damage in Rubizhne, Ukraine. True Positive and True Negative predictions are shown in red and green, respectively, while False Positives and False Negatives are shown in orange and purple. Percentages in the legend reflect the proportion of building area in each category. False positives are largely co-located with true positives.

\par}
\end{minipage}
  \label{Rubizhne_footprints}
\end{figure*}


The Area Under the Curve is insensitive to changes in the class distribution, and therefore provides a more comparable measure of discrimination across contexts. It is generated by calculating the true positive rate  and the false positive rate at all possible threshold values of T and plotting the resulting Receiver Operating Characteristic (ROC) curve, shown in Figure \ref{roc}. AUC of 0.5 suggests no discrimination, AUC of 1 indicates perfect discrimination, and in the field of Oncology, an AUC of “0.8 to 0.9 is considered excellent” \citep{mandrekar_receiver_2010}. The AUC across all of Ukraine is 0.88, and 0.81 for the Gaza Strip. AUC is lower for cities in Syria and Iraq, likely due to the fact that the reference periods used in the generation of the PWTT contained substantial fighting (and therefore, destruction), resulting in noisier estimates. 

Despite having the lowest AUC scores in the sample, the results for Syria rival-- and in some cases, outperform-- deep learning approaches to conflict damage detection. In their analysis of damage to six Syrian cities damaged during the civil war, Mueller et. al. (2021) employ 50cm resolution optical imagery, a convolutional neural network, and a second stage random forest model \citep{mueller_monitoring_2021}. They achieve AUC values ranging between 0.8 and 0.91 when training and testing on the same city, but when a model trained only Aleppo, AUC for Raqqa drops from 0.87 to 0.58, with similar reductions for other cities. In contrast, the PWTT manages to achieve an AUC score of 0.77 for Raqqa despite being unsupervised and using lower resolution open-access satellite imagery \citep{mueller_monitoring_2021}.

Though the PWTT enables robust discrimination between damaged and undamaged buildings, precision is consistently lower than recall for most cities. This suggests that false positives are more common than false negatives, even when selecting an optimal threshold value for damage classification. An analysis of false positives suggests that the majority thereof are driven by spillover effects of actual damage.

\subsection{Damage Spillovers}

Figure \ref{Rubizhne_footprints} plots predicted versus observed damage at the building level in Rubizhne, Ukraine. The vast majority of false positives (shown in orange) are in close proximity to true positives (shown in red). Across all cities, the median distance between a false positive and the nearest damaged building is just 31 meters, and 27\% of false positives are direct neighbours of a damaged building (within 10 meters of a damage label). In contrast, the median distance between a true negative and the nearest positive label is nearly six times greater, at 175 meters. The difference is highly statistically significant (t=146.1). Figure \ref{prox_hist} in the Appendix plots the histogram of distances between false positive and true negative predictions to the nearest damaged building.

There are a number of possible reasons for these spillovers. Given that roughly a third of false positives are less than one Sentinel-1 pixel away from a damaged building, spillovers are likely partially driven by resolution-related limitations in the process of precisely delineating between damaged and undamaged buildings. There may also be actual changes on the ground in the areas surrounding a damaged building such as rubble dispersed by the destructive event, or fires impacting nearby vegetation. It is also likely that buildings neighbouring those that have been destroyed have actually sustained some lateral damage that was missed by the manual damage annotation process but picked up by the side-looking SAR sensor. Thus, the PWTT correctly localizes damage within a city, as the majority of false positives generated by the PWTT algorithm result from a slight ($\sim$30 meter) over-prediction of the spatial extent of genuinely damaged areas. This is less of a threat to model utility than if the model falsely predicted entire swathes of a city to be damaged. 

Beyond the damage spillovers, there are several reasons for which perfect agreement between the damage predictions and the labeled building footprints cannot be achieved. The first set of challenges involves the labeled data itself. False negatives could result from a degree of human error in the process of labeling damaged buildings, particularly in regard to lateral damage that might not be visible from a nadir view. The quality of the Microsoft Buildings dataset, while generally high, is also imperfect; the boundaries of buildings with irregular shapes are sometimes misspecified, and a large number of small geometries that are not buildings have been generated. Many of these were removed in the process of data cleaning, but some remain. 

Further false negatives may be attributable to temporal mismatch between the prediction window and the date of the manual labeling; UNOSAT damage labeling is mostly conducted on the basis of a single high resolution satellite image, but the PWTT algorithm utilizes a post-conflict time window of two months following the date of the UNOSAT assessment. Damage occurring during this time window would register in the damage predictions, but would not be recorded in the UNOSAT data.

Finally, the resolution of the satellite imagery used in this analysis is a limiting factor when it comes to the identification of damage to small buildings in particular. Indeed, 26\% of the buildings in this sample are smaller than a single pixel in the Sentinel imagery, and 62\% are smaller than two pixels. Despite these constraints, the PWTT manages to generate building damage assessments that rival the accuracy of deep learning approaches.

\subsection{Damage Intensity}

While the identification of individual damaged buildings is an important output for practitioners, zonal damage assessments are also common; UNOSAT conducts Rapid Damage Assessments in which a 500m-by-500m grid is drawn over a city, and a grid-cell is labeled as damaged if there is at least one damaged building therein \citep{unosat_mariupol_2022}. The PWTT performs even better at this task; the binary classification of 500m grid-cells containing at least one damaged building yields an F1 accuracy of 0.79 using a single cutoff value of T across all cities, with an overall AUC of 0.81. Table S4 reports city-level accuracy statistics. However, with such large cells there is a high likelihood of observing multiple damaged buildings within the same cell; some grid-cells in Gaza contain nearly 400 damaged buildings. 

\begin{figure}[!h]
  \caption{Predicted and Observed Damage Intensity, Gaza}
      \begin{minipage}{\columnwidth}
  \includegraphics[width=\columnwidth]{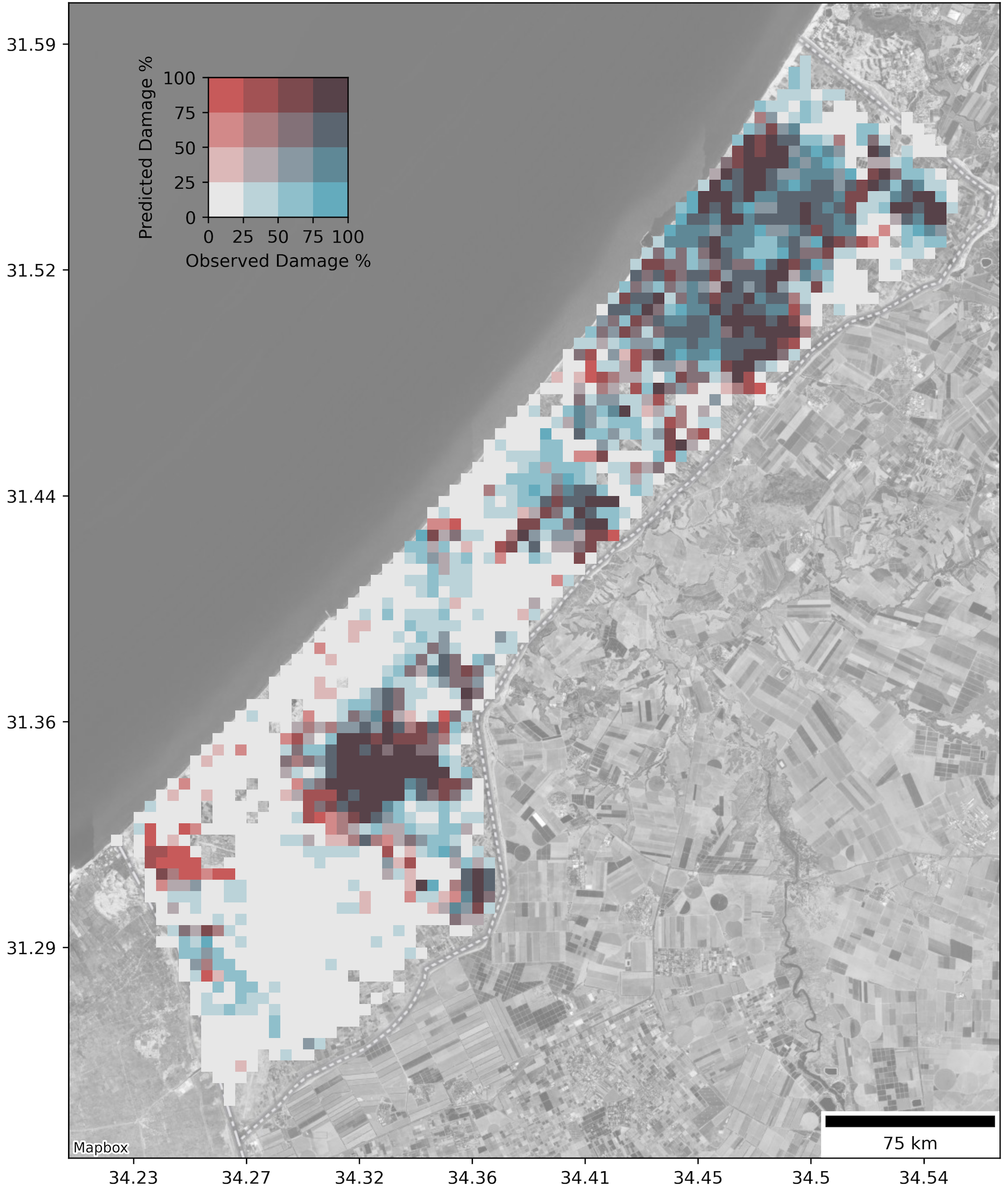}
{\scriptsize Purple cells indicate cells with both high predicted and observed damage, while white cells indicate low levels of both variables. Blue cells indicate an underprediction of damage intensity, while red cells indicate overprediction. “Observed Damage” refers to the number of damaged buildings in a grid cell, while “Predicted Damage” reflects the average T-value from the PWTT. The percentile distribution is used for both variables. \par}
\end{minipage}
  \label{Gaza_500m}
\end{figure}

Figure \ref{Gaza_500m} shows the spatial relationship between predicted damage (the mean value of T in the cell) and observed damage intensity (the number of damage labels in the cell) at the level of 500m grid cells in Gaza. There is strong agreement between the two, with a slight tendency to underpredict sparse damage and damage on the periphery of severely damaged areas. Given that both the cell-level average T-value and the number of damaged buildings are both scalar values, it would be more informative to frame zonal damage assessments as a regression problem rather than a binary classification problem with the following form: 

\begin{equation}
\log(D_{ic})=\beta_0+\beta_1 T_{ic} +\beta_2 B_ic+ \alpha_{l}+ \epsilon    
\end{equation}

\begin{table*}[h!]
\begin{center}

\begin{threeparttable}
\caption{Regression Results Between log Number of Damaged Buildings and Mean T-value, 500m Grid-Level}
\begin{tabular}{llllll}
\toprule
\midrule
               & Palestine & Ukraine   & Syria     & Iraq      & All       \\
\midrule
Intercept      & -0.907*** & -0.668*** & -0.582*** & -0.767*** & 0.073*    \\
               & (0.092)   & (0.038)   & (0.096)   & (0.084)   & (0.039)   \\
Mean T-Value       & 0.724***  & 0.345***  & 0.660***  & 0.529***  & 0.524***  \\
               & (0.029)   & (0.013)   & (0.041)   & (0.025)   & (0.011)   \\
Building Count       & 0.008***  & 0.006***  & 0.014***  & 0.005***  & 0.008***  \\
               & (0.000)   & (0.000)   & (0.001)   & (0.000)   & (0.000)   \\
\midrule
R-squared      & 0.531     & 0.529     & 0.634     & 0.463     & 0.633     \\
N              & 1780      & 4977      & 1188      & 1141      & 9086      \\
\midrule
\bottomrule
\end{tabular}

\begin{tablenotes}
\scriptsize
\item
* p<0.05, ** p<0.01, *** p<0.001. Country-level regression results between the log number of damaged buildings in a 500m grid cell are reported above. All specifications include a city-level fixed effect. The rightmost column uses all available grid-cells. 
\end{tablenotes}
\label{reg_tab}
\end{threeparttable}
\end{center}

\end{table*}

Where $\log(D_{ic})$ is the log number of damaged buildings in grid cell $i$ in country $c$, $T$ is the mean value of the damage probability raster in the corresponding cell, and $\alpha_{l}$ is a city-level fixed effect. Table \ref{reg_tab} reports the results of this specification. 

In every country, there is a strong positive relationship between the log number of damaged buildings in a grid cell and the mean value of the damage probability raster. Across all 12 cities, a one-unit increase in a grid cell’s mean T-value is associated with a 69\% increase in the number of damaged buildings therein. All coefficients are significant at the 99\% confidence level, though there is some range in their magnitudes. These differences reflect the differing spatial patterns of damage across geographic contexts. The average number of damaged buildings per grid cell in Ukraine is 3, while in Gaza it is 35. 

R-squared values from the regression models demonstrate a strong association between the damage probability raster and observed damage intensity. Across all cities, over 63\% of the variation in the number of damaged buildings in a cell can be explained by the average value of the damage probability raster therein. 

Framing zonal battle damage assessment as a regression problem has a number of practical benefits for practitioners. Given that over half of the grid-cells in the sample contain at least one damaged building and that some contain nearly 400, binary damage classification only yields a limited amount of information. Nevertheless, the change detection algorithm elaborated herein achieves an F1 accuracy of 0.79 for this task. Beyond simply identifying damaged buildings, the PWTT can accurately gauge the intensity of damage within a 2.5 km$^2$ area across different countries. 

\section{Discussion}

The generation of public information regarding buildings damaged by conflict has become particularly important in the context of recent, highly destructive wars in Gaza and Ukraine. This paper develops the Pixelwise T-Test (PWTT), a new algorithm for open-access battle damage estimation that is accurate, lightweight, and generalizable. 

The accuracy of the PWTT rivals that of deep learning approaches, achieving a building-level Area Under the Curve of 0.88 across 8 Ukrainian cities, and 0.81 for Gaza. It achieves an F1 score of 0.79 for the detection of damage within a 500m grid cell, and can even generate accurate estimates of damage intensity: the mean value of T within a grid cell explains 63\% of the variation in the log number of damaged buildings therein. 

The PWTT addresses many of the problems associated with expense, coverage consistency, and domain shift that affect deep-learning based approaches to building damage detection. The cost of high-resolution satellite imagery required to run inference using a deep learning model would exceed \$13 million for a single collection of Ukraine \citep{aaas_high-resolution_2024}. Even if the imagery could be sourced, cloud cover poses a significant challenge given that the annual cloud fraction exceeds 60\% in many parts of Ukraine \citep{nasa_cloud_2024}. The imagery used by the PWTT is not only open access but can penetrate through cloud cover, guaranteeing consistent coverage at a high repeat rate. Furthermore, while neural networks struggle to generalize even within the same country, the PWTT conducts unsupervised damage detection by incorporating a year’s worth of historical imagery for each pixel. 

These properties enable the creation of interactive Battle Damage Dashboards for \href{https://ollielballinger.users.earthengine.app/view/ukraine-damage-assessment}{Ukraine} and \href{https://ee-ollielballinger.projects.earthengine.app/view/gazadamage}{Gaza} that conduct multi-temporal, wide area building damage assessments on the fly using the Google Earth Engine cloud computing platform. These dashboards have three key features: First, they allow for damage assessments to be carried out across an entire country and for any time period during a war. Second, damage estimates are combined with high resolution pre-war population data to estimate the number of people previously living in now-destroyed areas. Third, they integrate additional open source data for additional verification in the form of geolocated social media footage of battle damage. Given the contentious nature of battle damage estimates, the use of open-source code and data ensures transparency, explainability, and replicability. Further details on the dashboards are available in the Appendix.  

The framework of the Battle Damage Dashboards is highly extensible, supporting modification to fit the needs of stakeholders. With light modification, alerts could be set up to detect the spread of fighting to different regions in near-real time, and provide early warning to nearby civilians. In a post-war reconstruction scenario, summaries of damage levels by administrative region could be easily generated to target field surveys and ultimate reconstruction efforts. 

There are nonetheless a number of limitations to the approach elaborated herein. Though the accuracy of the PWTT is assessed using over half a million building footprints in different climatic regions, further research must test the algorithm in Africa, Asia, and Latin America. A second area for improvement would be to increase accuracy by testing different change detection algorithms; Bayesian and non-parametric methods in particular hold promise. The creation of an open-access benchmark dataset for battle damage detection comprising over half a million labeled building footprints across 12 cities in four countries provides a common reference for future studies of battle damage.

While maximizing predictive accuracy is a natural goal, it is also incumbent upon researchers to consider the goal of battle damage assessment, which is not to fully automate the process. Any model is subject to error, and for high-stakes decision-making around post-conflict reconstruction there must always be a human in the loop. Where more precise damage estimates are required, the PWTT can be deployed to “tip and cue” manual assessments or field surveys.

\twocolumn
\newpage
\bibliographystyle{unsrtnat}
\bibliography{references} 

\newpage
\onecolumn

\setcounter{table}{0}
\renewcommand{\thetable}{S\arabic{table}}
\setcounter{figure}{0}
\renewcommand{\thefigure}{S\arabic{figure}}

\section{Appendix}

\subsection{Battle Damage Dashboards}

\subsubsection{Damage estimates}

The core feature of the Battle Damage Dashboard is that it enables building damage assessment to be carried out across an entire country over the full duration of a conflict. A date slider allows users to conduct damage detection for any two-month period during the war, and damaged building footprints can be downloaded within a user-defined area. 

Building-level precision and recall are critical measures of model performance for humanitarian practitioners, as they provide highly interpretable measures of uncertainty. Country-wide precision-recall curves at the building level are used in the visualization and analysis of estimated damage. The value of T that maximizes F1 accuracy is used as the cutoff value for damage prediction in each country. Areas with lower T values are masked, and areas with higher values are colored using a color scale. Corresponding precision values are communicated for the color scale. For example in the Ukraine Battle Damage Dashboard, a yellow pixel has a 60\% chance of being damaged, while a purple area has an >83\% chance of being damaged. 

\subsubsection{Population Exposure}

Damage predictions are combined with population density estimates from Oak Ridge National Laboratory’s LandScan HD program to quantify the number of people likely to be permanently displaced by battle damage. LandScan data has been used extensively in the literature on disaster response to estimate populations affected by natural hazards and conflict \citep{laverdiere_landscan_2022, urban_towards_2023}. Ambient population distributions are estimated by combining high resolution building features, occupancy data, land use, and data on Points of Interest \citep{urban_towards_2023}. LandScanHD data are made available at a spatial resolution of 90 meters. Once damaged areas have been identified via the PWTT, the ambient pre-war population in the damaged area is calculated using LandScan data. Importantly, these data reflect the pre-war population distribution and as such they reflect the number of people previously living in now-destroyed areas. 

This provides an important additional layer of information for humanitarian practitioners, enabling more precise needs assessments based on whether fighting is predominantly occurring in industrial zones, high-density urban areas, or suburban neighbourhoods. For example, the town of Popasna in Ukraine sustained heavy damage, resulting in over 1000 damaged buildings-- largely single family homes-- with a corresponding pre-war population of around 8000. An equivalent number of damaged buildings in Gaza City’s Rimal neighbourhood permanently displaces roughly 40000 people. The addition of population density data allows the Battle Damage Dashboards to overcome a key limitation of using the number, proportion, or total area of damaged buildings in humanitarian response. 

\subsubsection{Geolocated Social Media Footage}

While all accuracy statistics are calculated on the basis of UNOSAT damage labels, these are only available for a handful of cities. As such, an additional source of verification data is provided in the form of geolocated footage from social media showing battle damage. These are sourced from the Centre for Information Resilience, an organization which maintains a crowdsourced database of photos and videos of conflict events extracted from social media platforms such as Telegram, TikTok, and Twitter \citep{strick_eyes_2023}. The source media is archived, chrono- and geolocated, categorized, and finally reviewed by a team of investigators. In the battle damage dashboards, clicking on a geolocated event will open a panel showing a brief description of the event, the date, a link to the source media, and a link to the geolocation of the event. Despite the review process, it should be noted that geolocated photos and videos from social media are subject to some level of error and should not be treated as ground truth. Nevertheless, they provide an important additional source of corroborative evidence in areas where UNOSAT damage labels are missing.

\newpage
\begin{figure*}[hp]
  \caption{Ukraine Battle Damage Dashboard Analysis of Bakhmut}
      \begin{minipage}{\textwidth}
  \includegraphics[width=\textwidth]{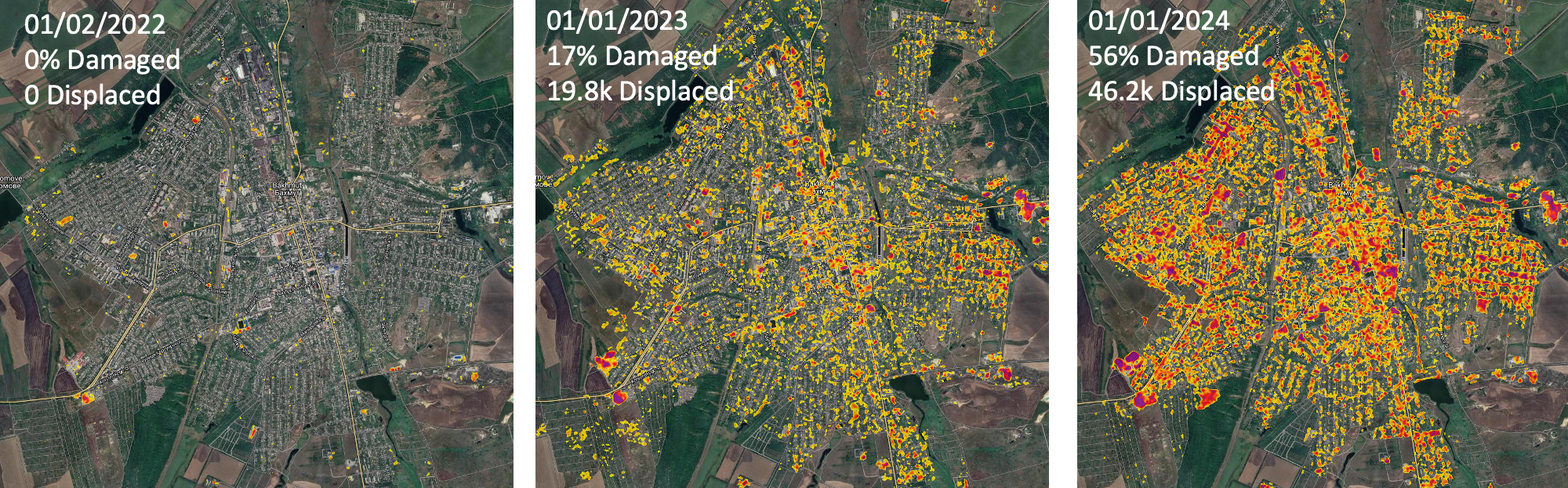}
{\scriptsize Building damage and population displacement estimates for Bakhmut generated at three time intervals for the city of Bakhmut, Ukraine, using the Battle Damage Dashboard. \par}
\end{minipage}
  \label{bakhmut}
\end{figure*}

\newpage

\newpage

\begin{table*}[hp]
\begin{center}

\begin{threeparttable}
\caption{Building-Level Damage Detection Accuracy Using a Balanced Sample}

\begin{tabular}{llllll}
\toprule
\hline
City & AUC & F1 & Precision & Recall & N \\

\midrule
Palestine & 0.82 & 0.76 & 0.67 & 0.89 & 136612 \\
Ukraine & 0.88 & 0.85 & 0.81 & 0.9 & 30150 \\
Syria & 0.73 & 0.75 & 0.65 & 0.89 & 62994 \\
Iraq & 0.77 & 0.74 & 0.67 & 0.81 & 31720 \\
\midrule
Gaza & 0.82 & 0.76 & 0.67 & 0.89 & 136612 \\
Lysychansk & 0.79 & 0.84 & 0.74 & 0.97 & 3006 \\
Sievierodonetsk & 0.72 & 0.82 & 0.71 & 0.98 & 2904 \\
Rubizhne & 0.8 & 0.77 & 0.67 & 0.9 & 5992 \\
Kharkiv & 0.83 & 0.87 & 0.84 & 0.91 & 1842 \\
Mariupol & 0.71 & 0.82 & 0.71 & 0.96 & 11592 \\
Irpin & 0.88 & 0.86 & 0.83 & 0.9 & 1634 \\
Hostomel & 0.82 & 0.87 & 0.85 & 0.89 & 1236 \\
Chernihiv & 0.89 & 0.87 & 0.88 & 0.87 & 1944 \\
Raqqa & 0.77 & 0.78 & 0.69 & 0.9 & 22398 \\
Mosul & 0.77 & 0.74 & 0.66 & 0.83 & 31720 \\
Aleppo & 0.71 & 0.75 & 0.65 & 0.88 & 40596 \\
\hline
All & 0.78 & 0.77 & 0.66 & 0.93 & 261476 \\
\hline
\bottomrule
\end{tabular}

\begin{tablenotes}
\scriptsize
\item
Building-level accuracy statistics are generated using UNOSAT damage annotations joined with building footprints. For each location, a threshold value of T is selected using a precision-recall curve to generate a binary damage classification. All statistics are weighted by building area. Balanced samples were generated by randomly sampling the same number of undamaged
buildings as damaged buildings for each country

\end{tablenotes}
\label{bal_results_tab}
\end{threeparttable}
\end{center}
\end{table*}

\newpage
\begin{table*}[hp]
\begin{center}
\begin{threeparttable}
\caption{500m Grid-Cell Level Damage Detection Accuracy}
\begin{tabular}{llllll}
\toprule
\midrule
Location & AUC & F1 & Precision & Recall & N \\
\midrule
Gaza & 0.79 & 0.88 & 0.79 & 0.99 & 1780 \\
Lysychansk & 0.71 & 0.85 & 0.75 & 0.98 & 419 \\
Sievierodonetsk & 0.79 & 0.88 & 0.83 & 0.94 & 253 \\
Rubizhne & 0.87 & 0.87 & 0.81 & 0.93 & 217 \\
Kharkiv & 0.72 & 0.42 & 0.36 & 0.51 & 2588 \\
Mariupol & 0.81 & 0.94 & 0.9 & 0.98 & 347 \\
Irpin & 0.86 & 0.88 & 0.8 & 0.99 & 226 \\
Hostomel & 0.74 & 0.83 & 0.72 & 0.98 & 186 \\
Chernihiv & 0.73 & 0.62 & 0.53 & 0.74 & 741 \\
Raqqa & 0.93 & 0.96 & 0.93 & 0.99 & 195 \\
Mosul & 0.74 & 0.88 & 0.8 & 0.99 & 1141 \\
Aleppo & 0.71 & 0.88 & 0.79 & 1.0 & 993 \\
\midrule
All & 0.81 & 0.79 & 0.78 & 0.8 & 9086 \\
\midrule
\bottomrule
\end{tabular}

\begin{tablenotes}
\scriptsize
\item
500m-by-500m grid-cell level accuracy statistics are generated using UNOSAT damage annotations. For each city, a threshold value of T is selected using a precision-recall curve to generate a binary damage classification. A cell is labeled damaged if it contains one or more damage annotations. 
\end{tablenotes}
\label{500m_tab}
\end{threeparttable}
\end{center}
\end{table*}


\newpage
\begin{figure*}[hp]
  \caption{Reference and inference periods for three cities}
      \begin{centering}
      \begin{minipage}{\textwidth}
  \includegraphics[width=0.9\textwidth]{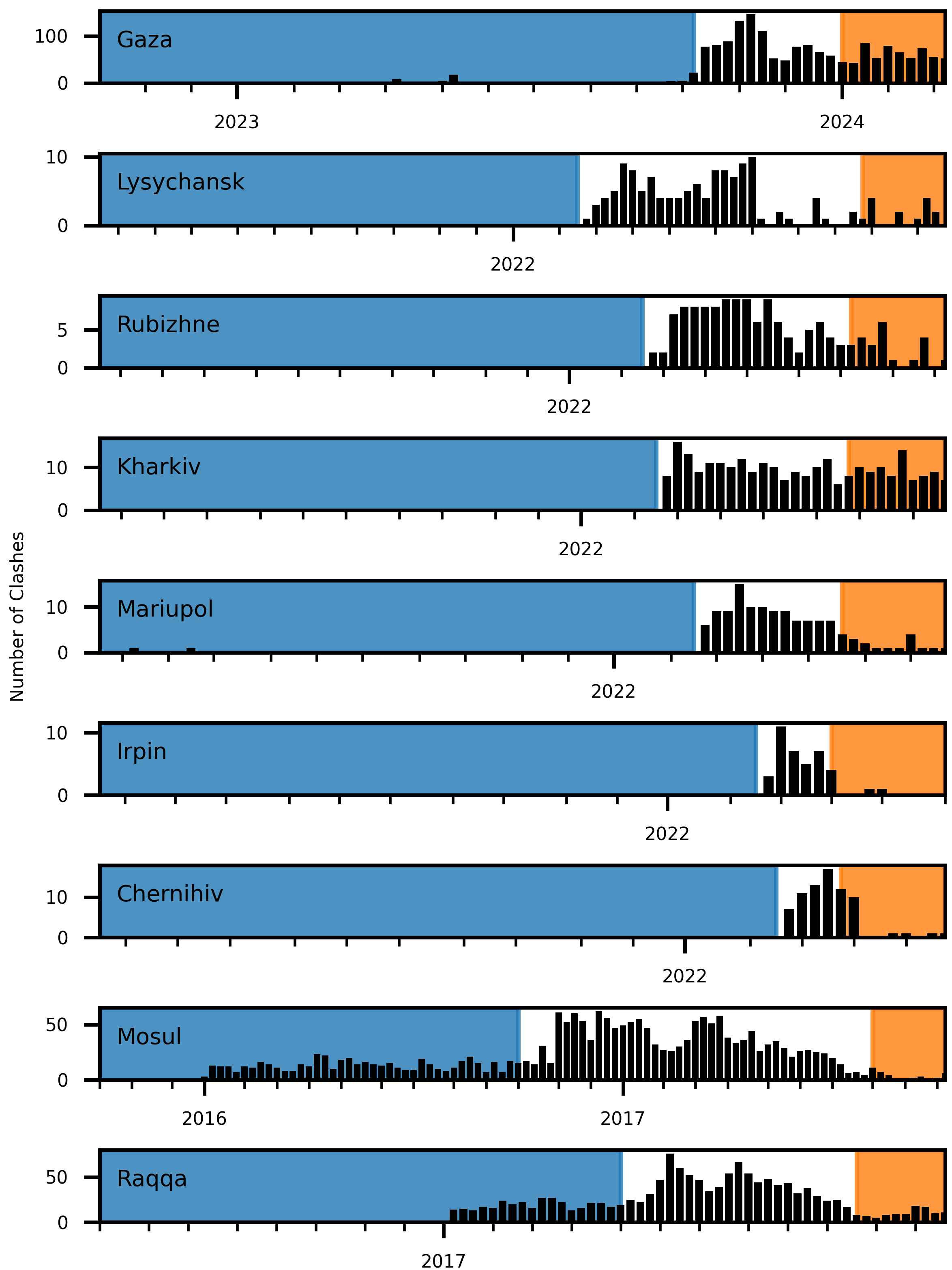}
{\scriptsize 
\\
The bars represent the weekly number of clashes in each city, sourced from the Armed Conflict Location Event (ACLED) dataset. 12 month pre-war reference periods are shaded in blue, spanning one year before the onset of conflict. 2 month post-war inference periods are shaded in orange, beginning with the date of the UNOSAT damage assessment.  
\par}
\end{minipage}
\end{centering}
  \label{acled}
\end{figure*}

\newpage

\begin{figure*}[hp]
  \caption{Distance between False Positives and True Negatives to the nearest Damaged Building}
      \begin{minipage}{\columnwidth}
  \includegraphics[width=\columnwidth]{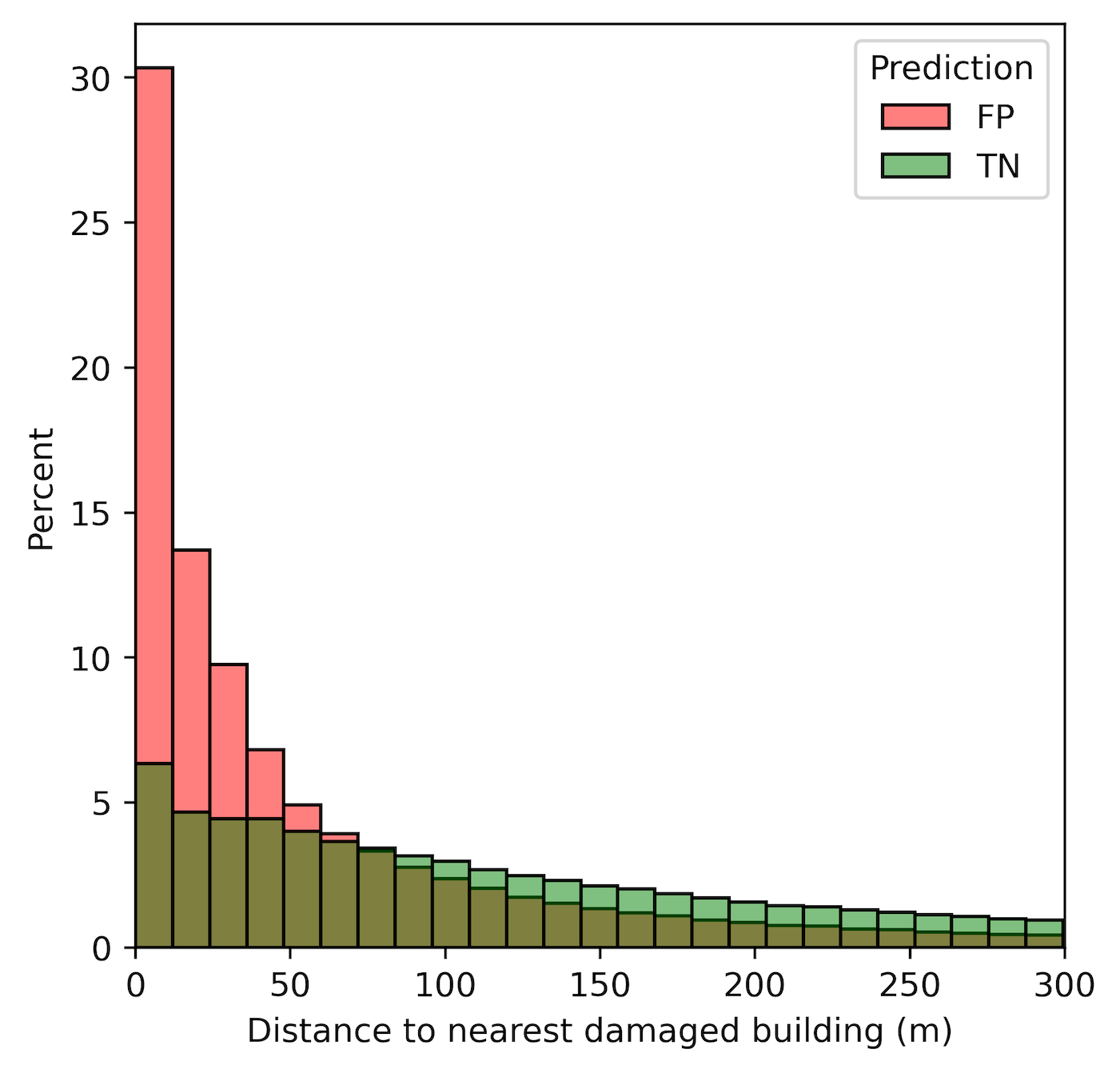}
{\scriptsize 
The distance between each False Positive (FP) prediction and the nearest UNOSAT damage annotation is calculated, and the resulting histogram is plotted in red. The same procedure is conducted for True Negative (TN) predictions, and plotted in green. False positives tend to be much closer to damaged buildings than True Negatives, suggesting spillover effects from damage. 
\par}
\end{minipage}
  \label{prox_hist}
\end{figure*}

\newpage
\begin{figure*}[hp]
  \caption{City-level ROC and Precision-Recall Curves}
      \begin{minipage}{\textwidth}
  \includegraphics[width=\textwidth]{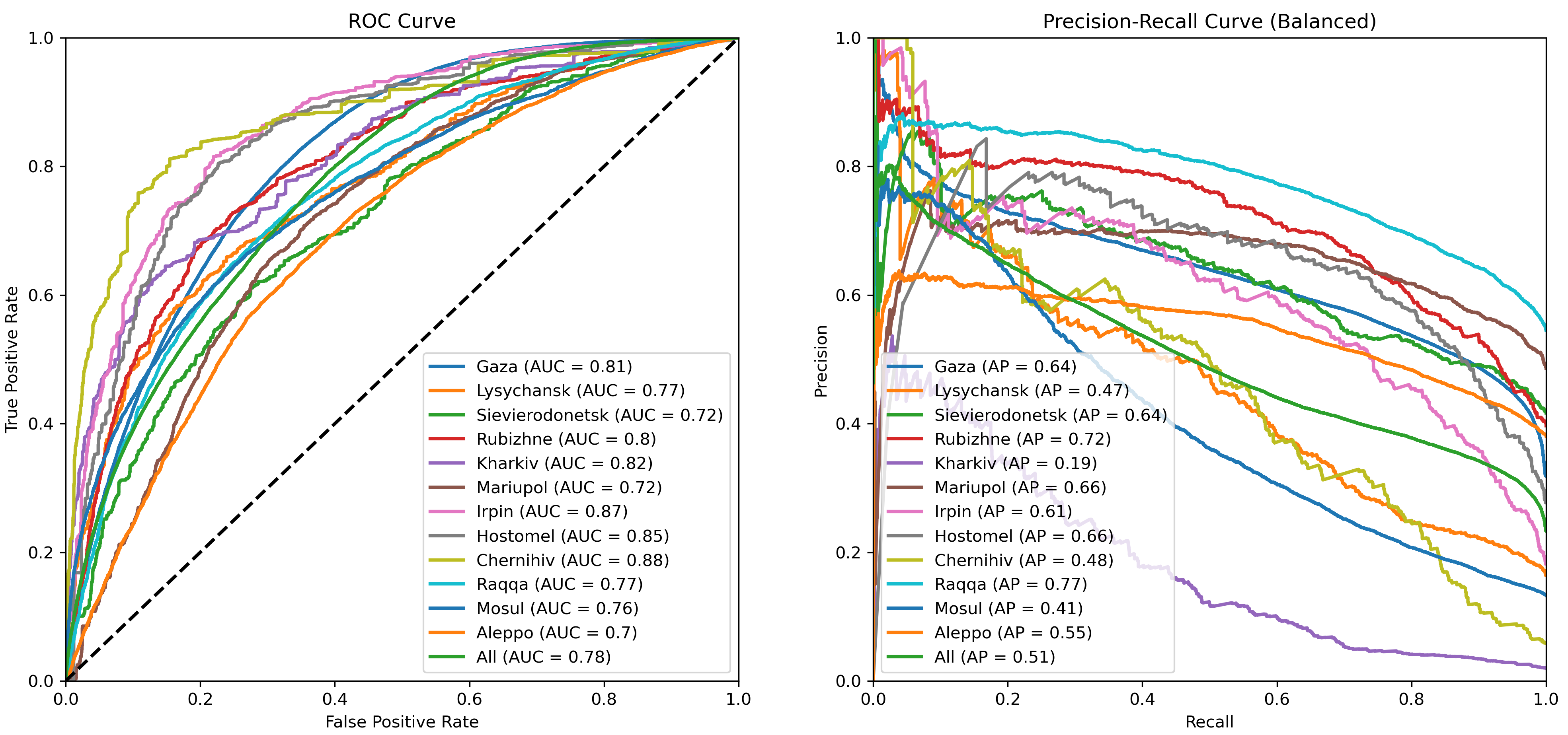}
{\scriptsize 
The Receiver Operating Characteristic (ROC) curve plots the true positive and false positive rates at each threshold value of T. The Precision-Recall curves does the same with precision and recall, using a balanced sample. Balanced samples were generated by randomly sampling the same number of undamaged buildings as damaged buildings for each city. 
\par}
\end{minipage}
  \label{roc_city}
\end{figure*}

\newpage
\begin{figure*}[hp]
  \caption{Predicted and Observed Damage, Irpin}
      \begin{minipage}{\textwidth}
  \includegraphics[width=\textwidth]{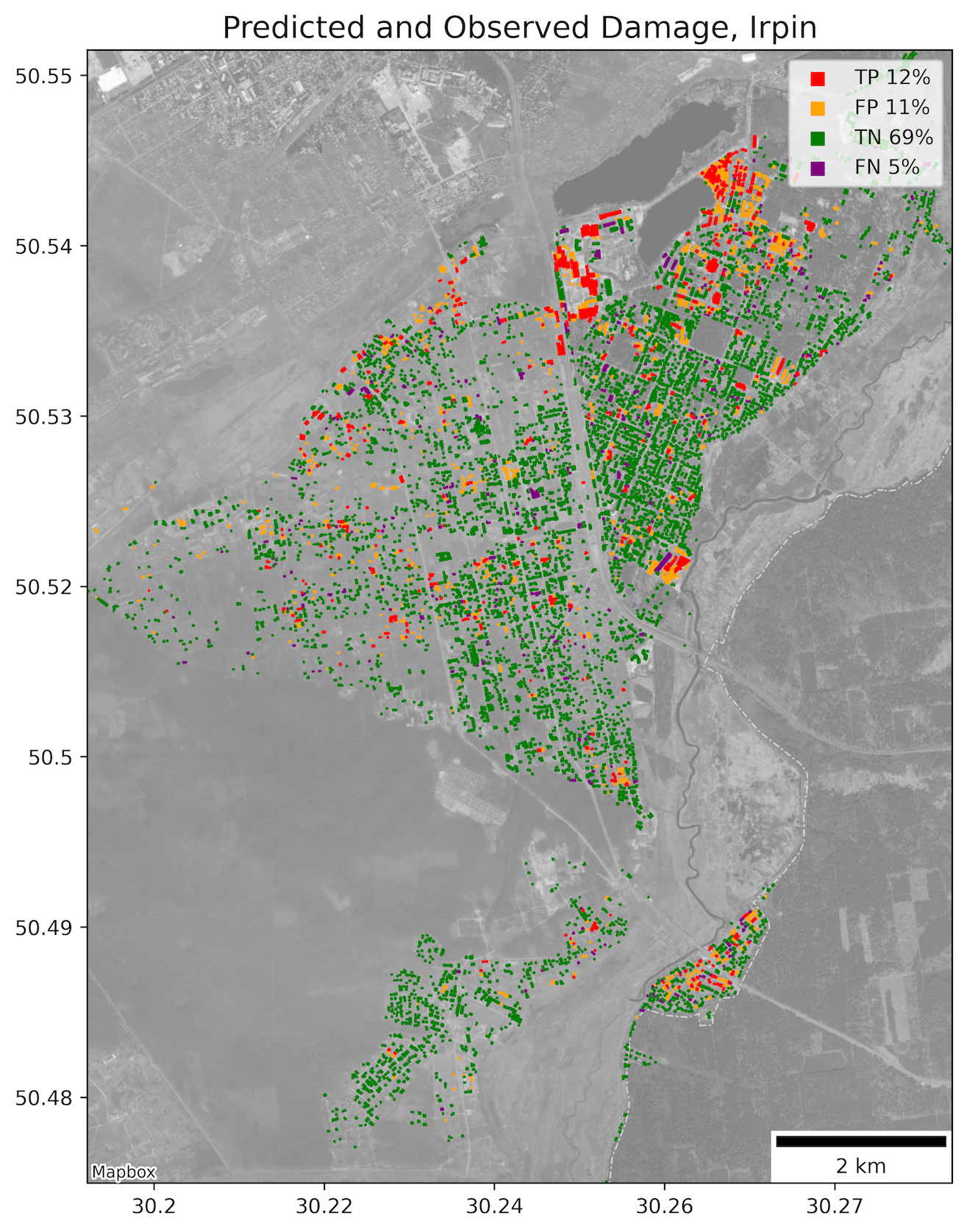}
{\scriptsize 
Predicted damage is compared to observed damage in Irpin, Ukraine. True Positive and True Negative predictions are shown in red and green, respectively, while False Positives and False Negatives are shown in orange and purple.
\par}
\end{minipage}
  \label{Irpin_footprints}
\end{figure*}

\newpage
\begin{figure*}[hp]
  \caption{Predicted and Observed Damage, Chernihiv}
      \begin{minipage}{\textwidth}
  \includegraphics[width=\textwidth]{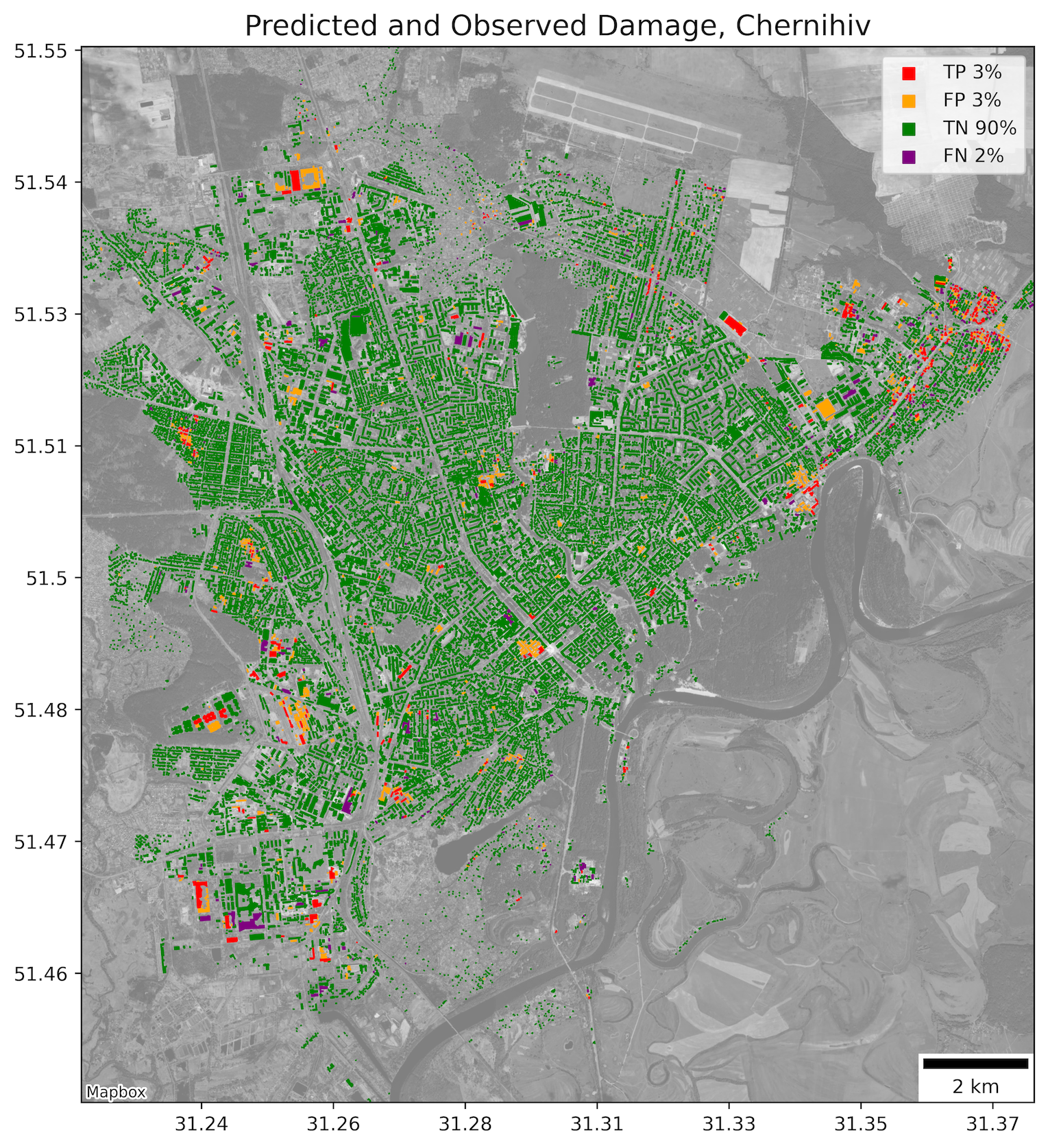}
{\scriptsize 
Predicted damage is compared to observed damage in Chernihiv, Ukraine. True Positive and True Negative predictions are shown in red and green, respectively, while False Positives and False Negatives are shown in orange and purple.
\par}
\end{minipage}
  \label{Chernihiv_footprints}
\end{figure*}

\newpage
\begin{figure*}[hp]
  \caption{Predicted and Observed Damage, Hostomel}
      \begin{minipage}{\textwidth}
  \includegraphics[width=\textwidth]{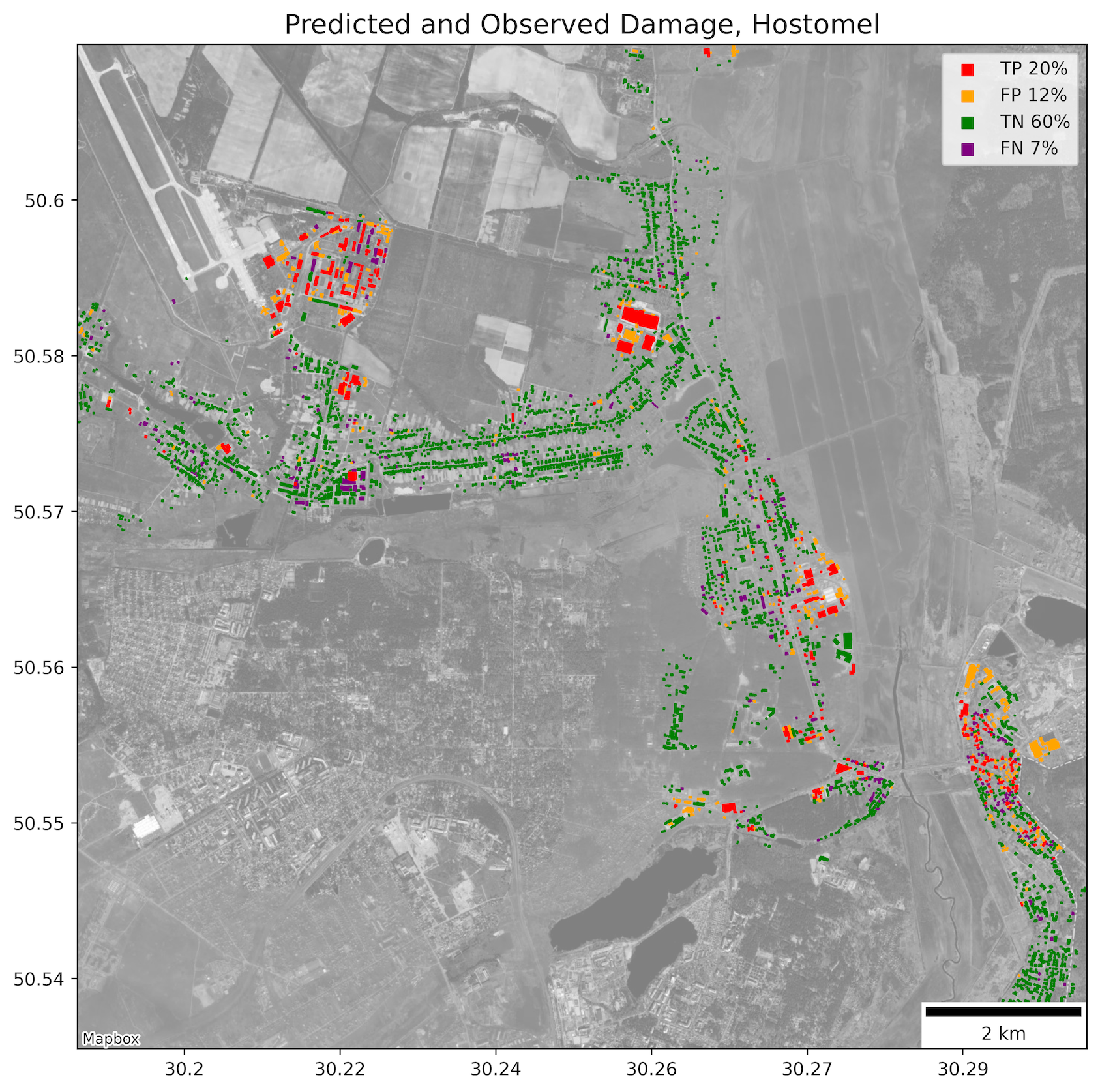}
{\scriptsize 
Predicted damage is compared to observed damage in Hostomel, Ukraine. True Positive and True Negative predictions are shown in red and green, respectively, while False Positives and False Negatives are shown in orange and purple.
\par}
\end{minipage}
  \label{Hostomel_footprints}
\end{figure*}

\newpage
\begin{figure*}[hp]
  \caption{Predicted and Observed Damage, Raqqa}
      \begin{minipage}{\textwidth}
  \includegraphics[width=\textwidth]{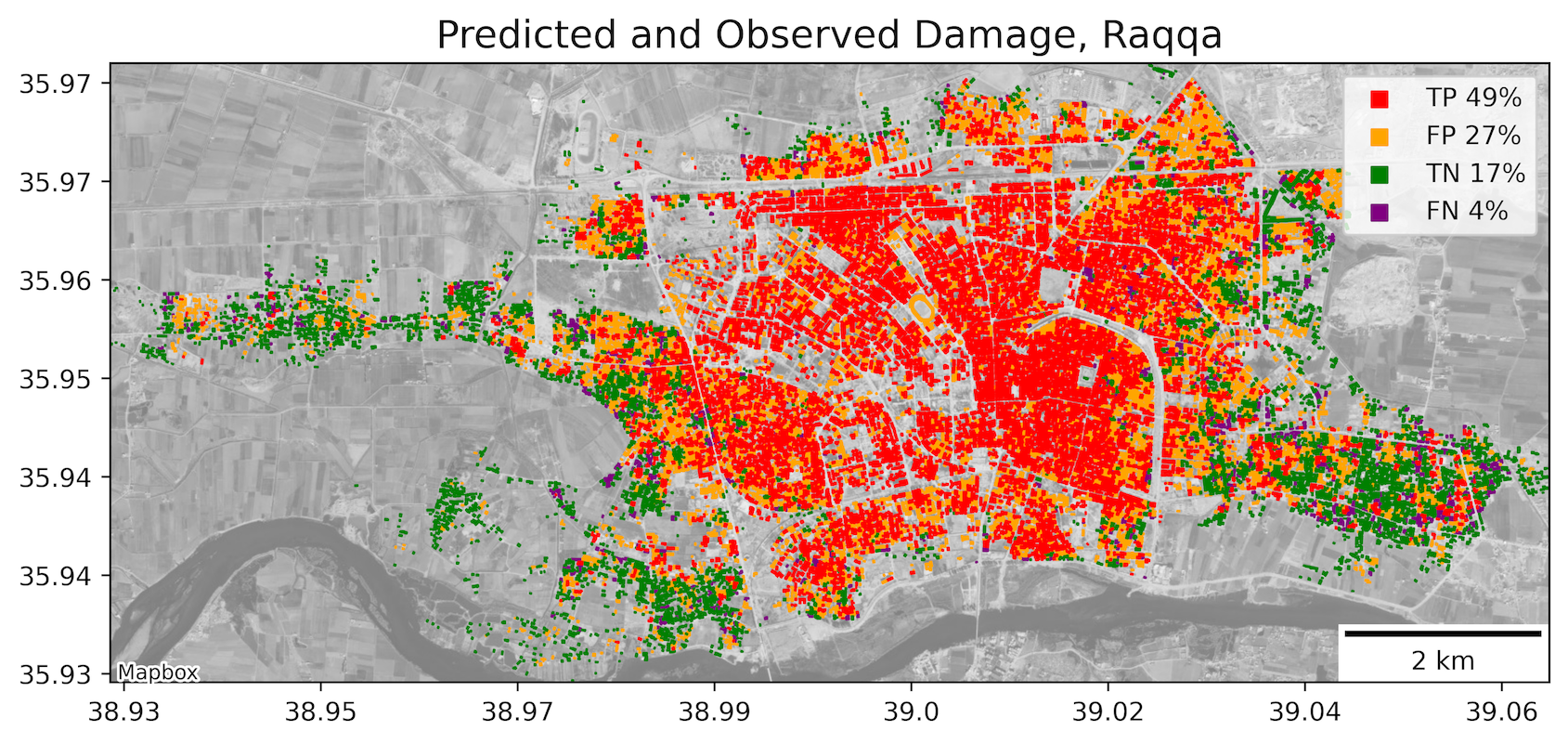}
{\scriptsize 
Predicted damage is compared to observed damage in Raqqa, Syria. True Positive and True Negative predictions are shown in red and green, respectively, while False Positives and False Negatives are shown in orange and purple.
\par}
\end{minipage}
  \label{Raqqa_footprints}
\end{figure*}


\end{document}